\documentclass[11pt]{article}

\usepackage[a4paper,margin=1in]{geometry}
\usepackage{times}
\usepackage{graphicx}
\usepackage{booktabs}
\usepackage{multirow}
\usepackage{array}
\usepackage{tabularx}
\usepackage{longtable}
\usepackage{float}
\usepackage{amsmath}
\usepackage{enumitem}
\usepackage[hidelinks]{hyperref}
\usepackage{xcolor}
\usepackage{amssymb}   % for \checkmark; use math \times
\newcommand{\modelname}{GSI Agent}
\newcommand{\basedmodel}{Qwen3-VL-2B-Instruct}
\newcommand{\gsidataset}{GSI Dataset}
\newcommand{\ckdataset}{Common Knowledge Dataset}

\title{\modelname: Domain Knowledge Enhancement for Large Language Models in Green Stormwater Infrastructure}
\author{Shaohuang Wang}
\date{}

\begin{document}
\maketitle

\begin{abstract}
Green Stormwater Infrastructure (GSI) systems, such as permeable pavement, rain gardens, and bioretention facilities, require continuous inspection and maintenance to ensure long-term performance. However, domain knowledge about GSI is often scattered across municipal manuals, regulatory documents, and inspection forms. As a result, non-expert users and maintenance staff may struggle to obtain reliable and actionable guidance from field observations. 
Although Large Language Models (LLMs) have demonstrated strong general reasoning and language generation capabilities, they often lack domain-specific knowledge and may produce inaccurate or hallucinated answers in engineering scenarios. This limitation restricts their direct application to professional infrastructure tasks.
In this paper, we propose \modelname{}, a domain-enhanced LLM framework designed to improve performance in GSI-related tasks. Our approach integrates three complementary strategies: (1) supervised fine-tuning (SFT) on a curated GSI instruction dataset, (2) retrieval-augmented generation (RAG) over an internal GSI knowledge base constructed from municipal documents, and (3) an agent-based reasoning pipeline that coordinates retrieval, context integration, and structured response generation. 
We also construct a new \gsidataset{} aligned with real-world GSI inspection and maintenance scenarios. Experimental results show that our framework significantly improves domain-specific performance while maintaining general knowledge capability. On the GSI dataset, BLEU-4 improves from 0.090 to 0.307, while performance on the common knowledge dataset remains stable (0.304 vs. 0.305). These results demonstrate that systematic domain knowledge enhancement can effectively adapt general-purpose LLMs to professional infrastructure applications.
\end{abstract}

\section{Introduction}

Green Stormwater Infrastructure (GSI) is widely adopted to reduce urban flooding, improve water quality, and support sustainable city development. Typical GSI facilities include permeable pavement, rain gardens, bioretention basins, and infiltration systems. However, the performance of these systems depends heavily on regular inspection and proper maintenance. Field staff, engineers, and community members often need clear and consistent guidance when identifying issues such as clogging, sediment accumulation, vegetation overgrowth, or structural damage.

In practice, GSI knowledge is distributed across multiple sources, including municipal design manuals, maintenance guidelines, regulatory documents, and inspection forms. These documents contain detailed procedural rules, technical constraints, and compliance requirements. However, the information is not centralized, and it can be difficult for non-expert users to quickly retrieve and interpret relevant guidance for specific field scenarios.
Large Language Models (LLMs) have shown strong performance in general question answering and reasoning tasks. However, when applied to specialized engineering domains, they often suffer from two major limitations. First, they lack up-to-date and document-grounded domain knowledge. Second, they may generate plausible but incorrect responses, especially when specific regulations or procedures are required. Therefore, directly applying a general-purpose LLM to GSI tasks may lead to unreliable outputs.

To address this gap, we focus on improving the domain capability of LLMs for GSI applications through a systematic knowledge enhancement framework. Instead of designing a new model architecture, we enhance a general LLM using three complementary components:

\begin{itemize}
    \item \textbf{Supervised Fine-Tuning (SFT):} We construct a domain-specific instruction dataset and fine-tune the base LLM to learn GSI terminology, reasoning patterns, and structured response formats.
    \item \textbf{Retrieval-Augmented Generation (RAG):} We build an internal GSI knowledge base from municipal manuals and regulatory documents. During inference, relevant passages are retrieved and provided as additional context to improve factual grounding.
    \item \textbf{Agent-Based Coordination:} We design an agent workflow that organizes retrieval, context integration, and response generation into a structured reasoning pipeline, improving consistency and reliability.
\end{itemize}

To support supervised training and evaluation, we construct the \gsidataset{}, which contains instruction-style samples covering question answering, verification, information extraction, and procedural generation tasks in GSI scenarios. We evaluate our framework on both the domain dataset and a general common knowledge dataset to examine two key aspects: (1) domain performance improvement and (2) general knowledge retention.
Experimental results show that our approach substantially improves performance on GSI tasks without degrading general capability. In particular, BLEU-4 on the GSI dataset increases from 0.090 to 0.307 after applying our domain enhancement framework, while performance on the common knowledge dataset remains stable.

In summary, this work demonstrates that combining supervised fine-tuning, retrieval augmentation, and agent-based coordination provides an effective and practical solution for adapting general-purpose LLMs to professional infrastructure domains. Our framework offers a systematic approach to domain knowledge enhancement that can be extended to other engineering and regulatory applications.

\begin{table}[H]
\centering
\small
\begin{tabularx}{\linewidth}{p{0.25\linewidth}p{0.20\linewidth}X}
\toprule
\textbf{Category} & \textbf{Typical method} & \textbf{Key idea and trade-off} \\
\midrule
Dynamic injection & RAG & Retrieve external documents at inference time; easy to update, but depends on retrieval quality. \\
Static injection & SFT & Put domain patterns into model parameters; strong for style/tasks, but harder to update. \\
Model adapter & LoRA & Train small rank adapters; efficient and low-cost, but capacity is limited. \\
Prompt optimization & Prompt & Control behavior without changing weights; fast but may be brittle. \\
\bottomrule
\end{tabularx}
\caption{Knowledge injection taxonomy used in this paper.}
\end{table}

\section{Related Work}

\subsection{Domain-Specific Knowledge Injection}

Injecting domain-specific knowledge into large language models has become a critical research area for improving task-specific performance. Song et al.~\cite{song2025survey} provide a comprehensive survey categorizing knowledge injection methods into dynamic integration, static fine-tuning, parameter-efficient adaptation, and prompt-based guidance. Dynamic integration connects the model to external knowledge sources during inference, enabling access to up-to-date documents without retraining, which has been effective in domains such as legal and biomedical text. Static fine-tuning, or supervised fine-tuning (SFT), incorporates domain knowledge directly into model parameters by training on task-specific corpora; prior studies have shown that SFT significantly improves reasoning and accuracy in specialized domains, but updating the knowledge requires retraining. Parameter-efficient methods, such as LoRA~\cite{hu2021lora}, reduce computational costs by training only additional low-rank matrices while freezing the original weights, achieving adaptation with minimal resources. Prompt-based guidance adjusts model behavior through structured instructions or prompts, which is lightweight but less effective for complex professional tasks. In GSI and infrastructure-related domains, where the knowledge is detailed and highly structured, combining SFT, retrieval grounding, and agent orchestration is essential for achieving both accuracy and adaptability.

\subsection{Retrieval-Augmented Generation}

Retrieval-Augmented Generation (RAG) has been proposed to address the limitations of fixed-parameter models in knowledge-intensive tasks by combining LLMs with external information retrieval~\cite{lewis2020rag}. In this framework, relevant documents or passages are retrieved from a structured corpus based on input queries, and the retrieved content is incorporated into the model context to improve factual correctness and reduce hallucination. Subsequent research has refined retrieval strategies using dense vector representations, contrastive learning, and multi-hop reasoning, significantly improving performance in open-domain and domain-specific applications. RAG is particularly valuable in professional domains where regulations, manuals, and standards are lengthy and constantly updated, as it allows the model to access authoritative knowledge without full retraining. In infrastructure management, including GSI inspection and maintenance, retrieval grounding ensures that model outputs reflect current municipal guidelines, safety protocols, and operational procedures, which are difficult to embed fully in model parameters.

\subsection{Domain-Specific LLM Agents}

Large language model agents extend the reasoning capabilities of conventional LLMs by integrating planning, tool invocation, and structured task execution. Early studies, such as chain-of-thought prompting~\cite{wei2022chainofthought}, demonstrated that decomposing reasoning into explicit intermediate steps improves problem-solving quality. Building on this idea, the ReAct framework~\cite{yao2023react} combines reasoning and acting loops, allowing models to plan, query external tools, and revise actions iteratively. Recent surveys~\cite{luo2025survey} highlight that domain-specific agents often incorporate memory, workflow orchestration, and modular tool interactions, and they have been applied in scientific discovery~\cite{ren2025survey} and biomedical assistance~\cite{xu2025biomedical} to manage complex, multi-step tasks. Evaluation of agent performance emphasizes planning quality, tool-use accuracy, and long-horizon reasoning reliability~\cite{yehudai2025evaluation}. Unlike general-purpose agent studies, our approach focuses on infrastructure tasks, designing a GSI-specific LLM agent that integrates fine-tuned reasoning and retrieval-grounded knowledge into a structured generation pipeline suitable for inspection, maintenance, and compliance workflows.

\section{Methodology}
\begin{figure}[H]
  \centering
  \includegraphics[width=\linewidth]{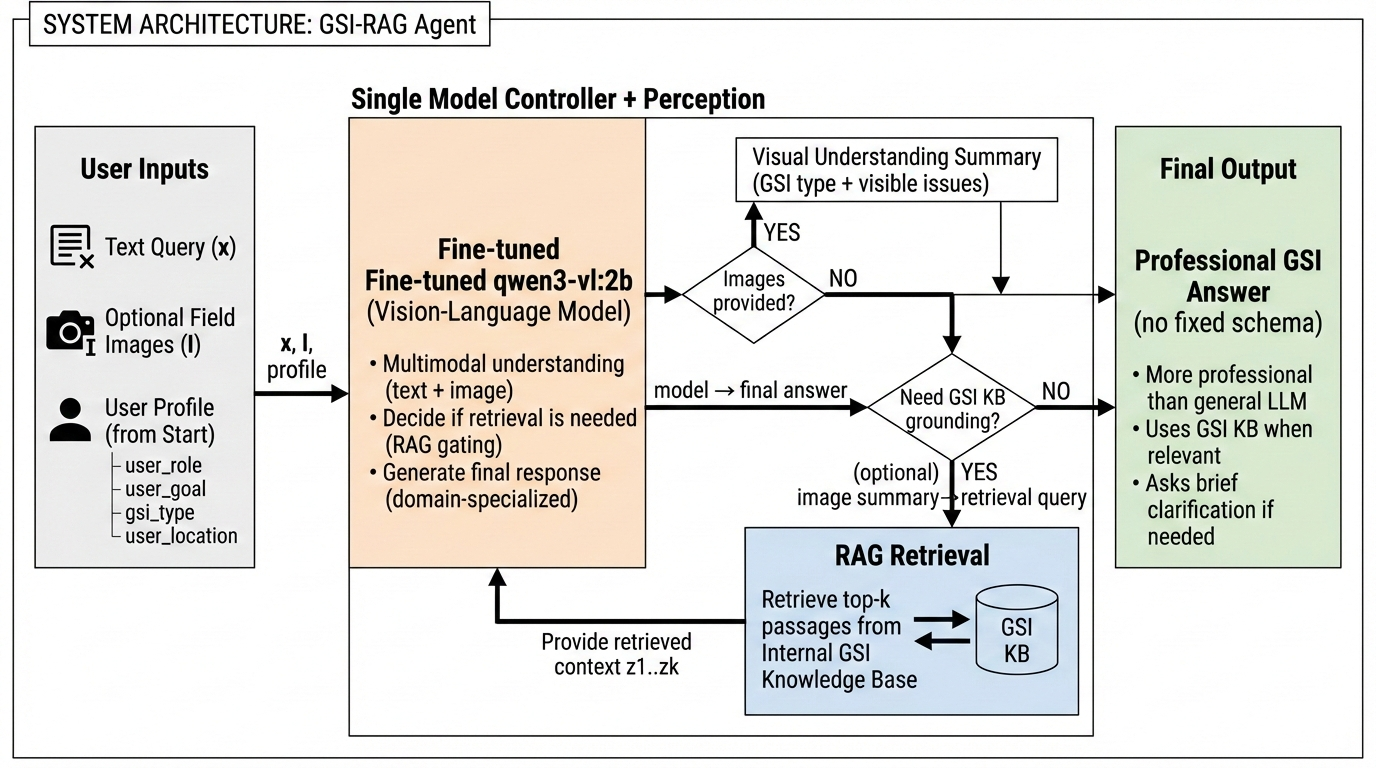}
  \caption{Architecture of the proposed GSI knowledge-enhanced LLM system integrating domain-specific fine-tuning, retrieval-augmented generation, and agent-based reasoning.}
  \label{fig:gsi_agent_arch}
\end{figure}

In this section, we present our approach to enhancing large language models (LLMs) for the Green Stormwater Infrastructure (GSI) domain. Our method integrates three complementary strategies: (i) domain-specific fine-tuning, (ii) retrieval-augmented generation (RAG), and (iii) domain-specific LLM agents. Figure~\ref{fig:gsi_agent_arch} illustrates the overall system architecture. The system is designed to provide professional, reliable, and context-aware responses to GSI-related queries by grounding outputs in verified domain knowledge. While the model can optionally process field images to support descriptive tasks, the core focus remains on textual reasoning and domain expertise.

\subsection{Domain-Specific Fine-Tuning}
To achieve domain expertise, we apply domain-specific fine-tuning on a general LLM using our curated GSI dataset (\gsidataset{}) following the methods summarized in EMNLP 2025 \cite{song2025survey}. We adopt parameter-efficient fine-tuning techniques (e.g., LoRA) to reduce computational cost while enabling the model to learn GSI-specific terminology, reasoning patterns, and regulatory context. This fine-tuning allows the LLM to interpret user queries accurately and generate responses aligned with engineering and planning practices. Optionally, the model can summarize field images to support descriptive assessments, such as identifying GSI types or visible maintenance issues, but this capability is secondary.

\subsection{Retrieval-Augmented Generation}
To improve factual accuracy and compliance with official guidelines, we incorporate a retrieval-augmented generation (RAG) pipeline. All GSI-related documents, including municipal manuals, inspection forms, and planning documents, are segmented into passages and embedded into a vector index. For each user query, the system retrieves the top-$k$ relevant passages and provides them as context to the fine-tuned LLM. This reduces hallucination, reinforces domain knowledge, and supports dynamic updates of the knowledge base without retraining the model. When field images are available, retrieval queries can optionally combine textual input and image summaries to enhance contextual relevance.

\subsection{Domain-Specific LLM Agents}
Finally, we implement domain-specific LLM agents to enable flexible, task-oriented reasoning. The agent combines the fine-tuned LLM with the RAG module and lightweight prompt control to support diverse GSI tasks, such as planning, inspection, and maintenance guidance. Rather than enforcing a rigid output format, the agent applies soft constraints: (i) utilize retrieved passages when relevant, (ii) avoid inventing regulations or technical standards, and (iii) ask concise clarification questions if critical information is missing. This design allows the system to adapt its responses to different user types—including engineers, planners, maintenance staff, and residents—while maintaining professional domain correctness.

%=======
%=======
%=======
%=======

\section{Experimental Setup}
In this section, we describe the datasets, evaluation metrics, baselines, and implementation details used to assess the effectiveness of our knowledge-enhanced LLM for GSI tasks. We aim to provide a comprehensive view of both domain-specific and general capabilities, supported by statistical summaries and visualizations.

\subsection{Datasets}
We evaluate our approach on two complementary datasets: a domain-specific GSI dataset (\gsidataset{}) and a general-purpose benchmark (\ckdataset{}). These datasets allow us to measure improvements in specialized GSI reasoning while monitoring the retention of broad LLM knowledge.

\subsubsection{\gsidataset{}}
\gsidataset{} is a curated, instruction-style dataset designed for supervised fine-tuning of GSI reasoning. It contains document-grounded examples covering diverse tasks such as question answering, verification, procedural guidance, and reasoning over regulatory standards. Each record includes explicit context, optional additional input, and a reference output grounded in official documents or field manuals (Table~\ref{tab:gsidataset-schema}).

\begin{table}[H]
\centering
\small
\begin{tabularx}{\linewidth}{p{0.25\linewidth}X}
\toprule
\textbf{Field} & \textbf{Description} \\
\midrule
\texttt{\_id} & Unique identifier (UUIDv4). \\
\texttt{\_source} & Source document (PDF) providing reference information. \\
\texttt{\_source\_location} & Geographical tag (e.g., ``Philadelphia, PA'') or empty if not location-specific. \\
\texttt{\_task\_type} & One of nine predefined task families. \\
\texttt{\_deployment\_type} & Intended usage: \texttt{fine-tuning} or \texttt{rag}. \\
\texttt{\_created\_at} & Timestamp of record creation (UTC, RFC3339). \\
instruction & Self-contained instruction with context for the model. \\
input & Optional supplemental context. \\
output & Reference answer grounded in official documents. \\
\bottomrule
\end{tabularx}
\caption{Schema of \gsidataset{} for instruction-based fine-tuning.}
\label{tab:gsidataset-schema}
\end{table}

The dataset contains 10,955 examples, with 54.2\% having a specific source location (e.g., Philadelphia) and 45.8\% location-agnostic. Deployment types are distributed as 73.3\% fine-tuning and 26.7\% retrieval-augmented generation (RAG) samples (Tables~\ref{tab:source-location} and \ref{tab:deployment-type}). The task family distribution reflects diverse reasoning capabilities, with the top three categories—question answering (31.2\%), verification/judgment (30.9\%), and generation/composition (15.2\%)—covering 77.3\% of the dataset (Table~\ref{tab:task-type}). These distributions indicate a balanced mix of knowledge-intensive and procedural reasoning tasks.

\begin{table}[H]
\centering
\caption{Source location distribution in \gsidataset{}.}
\label{tab:source-location}
\begin{tabular}{lrr}
\toprule
\textbf{Location} & \textbf{Count} & \textbf{Percentage} \\
\midrule
Philadelphia, PA & 5219 & 54.2\% \\
None             & 4791 & 45.8\% \\
\bottomrule
\end{tabular}
\end{table}

\begin{table}[H]
\centering
\caption{Deployment type distribution in \gsidataset{}.}
\label{tab:deployment-type}
\begin{tabular}{lrr}
\toprule
\textbf{Type} & \textbf{Count} & \textbf{Percentage} \\
\midrule
Fine-tuning & 7460 & 73.3\% \\
RAG         & 3495  & 26.7\% \\
\bottomrule
\end{tabular}
\end{table}

\begin{table}[H]
\centering
\caption{Task type distribution in \gsidataset{}.}
\label{tab:task-type}
\begin{tabular}{lrr}
\toprule
\textbf{Task Type} & \textbf{Count} & \textbf{Percentage} \\
\midrule
Question Answering        & 5300 & 31.2\% \\
Verification / Judgment   & 5241 & 30.9\% \\
Generation / Composition  & 2573 & 15.2\% \\
Information Extraction    & 1724 & 10.1\% \\
Classification            & 1225 & 7.2\%  \\
Reasoning / Math / Logic  & 758  & 4.4\%  \\
Dialogue Interaction      & 100  & 0.6\%  \\
Rewriting / Transformation& 41   & 0.2\%  \\
Code / Program Execution  & 0    & 0\%  \\
\bottomrule
\end{tabular}
\end{table}

\begin{figure}[H]
\centering
\includegraphics[width=0.7\linewidth]{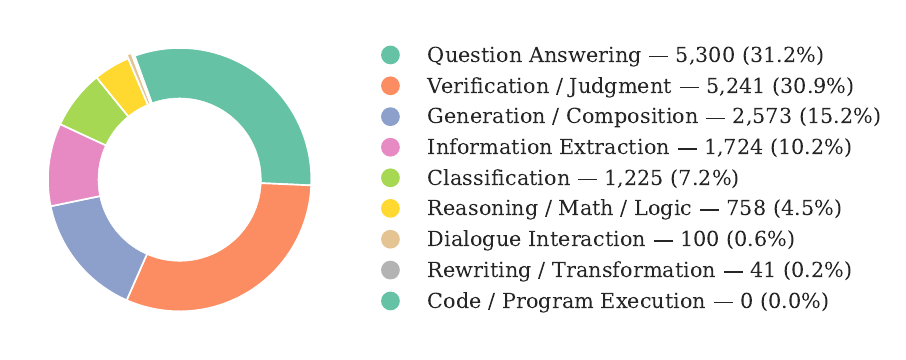}
\caption{Task type distribution in \gsidataset{}.}
\label{fig:gsidataset-distribution}
\end{figure}

Figure~\ref{fig:gsidataset-distribution} visualizes the task type distribution, highlighting the dominance of question-answering and verification tasks while maintaining coverage of procedural and reasoning-intensive examples.

\subsubsection{\ckdataset{}}
\ckdataset{} is a general benchmark used to measure knowledge retention outside the GSI domain. For our experiments, we sample 5,000 examples from publicly available LLM evaluation datasets such as MMMU/MMBench. This dataset includes question answering, classification, and reasoning tasks in diverse domains. Table~\ref{tab:ckdataset-summary} summarizes its characteristics. Visualizations of task-type distribution can be included similarly to \gsidataset{} for comparison.

\begin{table}[H]
\centering
\small
\begin{tabular}{lrr}
\toprule
\textbf{Task Type} & \textbf{Count} & \textbf{Percentage} \\
\midrule
Question Answering & 400 & 40\% \\
Classification     & 300 & 30\% \\
Reasoning / Logic  & 300 & 30\% \\
\bottomrule
\end{tabular}
\caption{Summary of \ckdataset{} used to evaluate general knowledge retention.}
\label{tab:ckdataset-summary}
\end{table}

By analyzing both datasets, we ensure that our LLM enhancements improve domain-specific reasoning without compromising general-purpose capabilities. Visual summaries and task statistics provide transparency and facilitate reproducibility.
%=======
%=======
%=======
%=======
%=======

\subsection{Metrics}
We evaluate outputs at multiple levels: lexical overlap, semantic similarity, judge-based quality, and human experts.
Table~\ref{tab:metrics} summarizes all metrics, and we give formal definitions below.

\begin{table}[H]
\centering
\small
\begin{tabularx}{\linewidth}{p{0.20\linewidth}p{0.18\linewidth}X}
\toprule
\textbf{Metric} & \textbf{Level} & \textbf{What it measures} \\
\midrule
BLEU-4 & Lexical & N-gram precision with brevity penalty; good for short factual text. \\
ROUGE-1/2/L & Lexical & Recall-style overlap; captures coverage for summary-like answers. \\
Micro-F1 & Label & Aggregated F1 for classification-style tasks (e.g., issue type). \\
Sentence-BERT & Semantic & Embedding cosine similarity; complements lexical overlap. \\
G-Eval (LLM Judge) & Semantic/Logic & LLM-based scoring for correctness and coherence. \\
Human Expert & Real & Expert rating on usefulness and correctness (small sample). \\
\bottomrule
\end{tabularx}
\caption{Evaluation metrics and their roles.}
\label{tab:metrics}
\end{table}

\paragraph{BLEU-4.}
BLEU-4 measures 1--4 gram precision with a brevity penalty to avoid overly short outputs:
\begin{equation}
\mathrm{BLEU}\text{-}4 = \mathrm{BP}\cdot \exp\left(\frac{1}{4}\sum_{n=1}^{4}\log p_n\right),
\end{equation}
where $p_n$ is the modified $n$-gram precision and $\mathrm{BP}$ penalizes too-short candidates.

\paragraph{ROUGE-1/2/L.}
ROUGE measures how much reference content is covered by the candidate (we report ROUGE-1, ROUGE-2, and ROUGE-L):
\begin{equation}
\mathrm{ROUGE}\text{-}n = \frac{\sum_{g\in \mathcal{G}_n(r)} \min(\mathrm{count}_c(g), \mathrm{count}_r(g))}
{\sum_{g\in \mathcal{G}_n(r)} \mathrm{count}_r(g)},
\end{equation}
where $\mathcal{G}_n(\cdot)$ is the multiset of $n$-grams (ROUGE-L is an LCS-based variant).

\paragraph{Micro-F1.}
For classification-style evaluation, we compute Micro-F1 by aggregating errors across all classes:
\begin{equation}
\mathrm{MicroF1}=\frac{2\,\mathrm{TP}}{2\,\mathrm{TP}+\mathrm{FP}+\mathrm{FN}},
\end{equation}
where $\mathrm{TP}$, $\mathrm{FP}$, and $\mathrm{FN}$ are total true positives, false positives, and false negatives.

\paragraph{Sentence-BERT similarity.}
We compute semantic similarity using cosine similarity between sentence embeddings:
\begin{equation}
\mathrm{SBERT}(c,r)=\frac{e_c^\top e_r}{\|e_c\|_2\ \|e_r\|_2},
\end{equation}
where $e_c$ and $e_r$ are embeddings of the candidate and reference answers.

\paragraph{G-Eval (LLM as a Judge).}
We ask an LLM judge to score each answer with a rubric-based score $s_i$ (e.g., 1--5) and report the mean:
\begin{equation}
\mathrm{G\text{-}Eval}=\frac{1}{N}\sum_{i=1}^{N} s_i.
\end{equation}
This metric helps evaluate correctness and coherence beyond surface similarity.

\paragraph{Human Expert.}
Similarly, human experts rate each answer (usefulness/correctness) and we report the average score:
\begin{equation}
\mathrm{HumanScore}=\frac{1}{N}\sum_{i=1}^{N} h_i,
\end{equation}
where $h_i$ is the expert score for sample $i$.

\subsection{Baselines}
In this section, we compare our proposed system against three baselines to quantify the effect of knowledge injection:

\begin{table}[H]
\centering
\small
\begin{tabularx}{\linewidth}{p{0.26\linewidth} c c c X}
\toprule
\textbf{Baseline} & \textbf{RAG} & \textbf{SFT} & \textbf{Agent} & \textbf{Notes} \\
\midrule
Base LLM & $\times$ & $\times$ & $\times$ & Direct prompting on base model. \\
Base LLM + RAG & \checkmark & $\times$ & $\times$ & Retrieval improves factuality; no parameter updates. \\
Fine-tuned LLM + RAG & \checkmark & \checkmark & \checkmark & Full system with LoRA-SFT, RAG, and agent reasoning. \\
\bottomrule
\end{tabularx}
\caption{Baselines used in evaluation.}
\label{tab:baselines}
\end{table}

We use \basedmodel{} as our primary base LLM and consider other open-source models (Qwen3-VL, InternVL, MiniCPM-V, Phi-3.5-Vision) mainly for feasibility comparison.

\subsection{Implementation Details}
In this section, we provide technical details of model fine-tuning. We adopt LoRA for parameter-efficient SFT on the Qwen3-VL instruction-tuned model. Table~\ref{tab:lora-config} summarizes the configuration.

\begin{table}[H]
\centering
\small
\begin{tabular}{lc}
\toprule
\textbf{Setting} & \textbf{Value} \\
\midrule
finetuning\_type & LoRA \\
bf16 & true \\
template & qwen3\_vl \\
lora\_alpha & 16 \\
lora\_dropout & 0 \\
lora\_rank & 8 \\
lora\_target & all \\
\bottomrule
\end{tabular}
\caption{LoRA fine-tuning configuration.}
\label{tab:lora-config}
\end{table}

\section{Experimental Results}
In this section, we report the main results and ablation studies, analyzing both general and domain-specific performance.

\subsection{Main Results}
Table~\ref{tab:main_results} presents evaluation results on \ckdataset{} and \gsidataset{}. We observe that our knowledge-enhanced LLM maintains general knowledge performance while achieving substantial improvement on GSI-specific tasks. BLEU-4 increases from 0.090 to 0.307 on \gsidataset{}, indicating strong domain adaptation. Sentence-BERT and G-Eval scores also show improved semantic correctness and reasoning quality.

\begin{table}[H]
\centering
\small
\begin{tabular}{lcccc}
\toprule
\multirow{2}{*}{\textbf{Metric}} &
\multicolumn{2}{c}{\textbf{\ckdataset{}}} &
\multicolumn{2}{c}{\textbf{\gsidataset{}}} \\
\cmidrule(lr){2-3}\cmidrule(lr){4-5}
& Base LLM & GSI LLM & Base LLM & GSI LLM \\
\midrule
BLEU-4 & 0.304 & 0.305 & 0.090 & 0.307 \\
ROUGE-1& 0.352 & 0.351 & 0.157 & 0.204 \\
ROUGE-2& 0.146 & 0.146 & 0.032 & 0.111 \\
ROUGE-L& 0.223 & 0.223 & 0.071 & 0.153 \\
Sentence-BERT& 0.861 & 0.869 & 0.544 & 0.742 \\
G-Eval & 0.82 & 0.84 & 0.57 & 0.79 \\
\bottomrule
\end{tabular}
\caption{Main results on general and domain datasets.}
\label{tab:main_results}
\end{table}

\subsection{Ablation Study}
To understand the contribution of each knowledge-injection strategy, we conduct an ablation study using G-Eval as the primary metric. We compare three settings: (i) LLM + RAG, (ii) LLM + Fine-tuning, and (iii) LLM + RAG + Fine-tuning. Table~\ref{tab:ablation} shows that the hybrid approach achieves the best balance between factual grounding and procedural reasoning, confirming that GSI tasks benefit from combining dynamic retrieval and learned domain capabilities.

\begin{table}[H]
\centering
\small
\begin{tabular}{lc}
\toprule
\textbf{Method} & \textbf{G-Eval Score} \\
\midrule
LLM + RAG & 0.51 \\
LLM + Fine-tuning & 0.63 \\
LLM + RAG + Fine-tuning & 0.72 \\
\bottomrule
\end{tabular}
\caption{Ablation study of knowledge-injection strategies.}
\label{tab:ablation}
\end{table}

\section{Conclusion}
In this section, we summarize our findings. We present \modelname{}, a knowledge-enhanced LLM system for GSI tasks, which maintains general knowledge performance while substantially improving domain-specific reasoning. Our experiments demonstrate that combining fine-tuning, retrieval-augmented generation, and agent-based reasoning yields the best overall performance. Future work includes scaling human expert evaluation, refining retrieval strategies, and performing finer-grained error analysis for real-world GSI maintenance applications.

\appendix

\section{Data Sources for RAG Corpus}
\label{app:sources}
We list the documents used to build the retrieval corpus. This table is designed to be extended: you can add a ``Used?'' column if some links are excluded.

\small
\begin{longtable}{p{0.05\linewidth}p{0.33\linewidth}p{0.10\linewidth}p{0.45\linewidth}}
\toprule
\textbf{\#} & \textbf{Title} & \textbf{Year} & \textbf{Link / Notes} \\
\midrule
\endfirsthead

\toprule
\textbf{\#} & \textbf{Title} & \textbf{Year} & \textbf{Link / Notes} \\
\midrule
\endhead

1 & Stormwater Management Guidance Manual & 2023 & \url{https://water.phila.gov/wp-content/uploads/files/stormwater-management-guidance-manual.pdf} \\
2 & Pennsylvania Stormwater BMPs Manual & -- & \url{https://greenport.pa.gov/elibrary/GetFolder?FolderID=1368916} \\
3 & City of Philadelphia Green Streets Design Manual & 2014 & \url{https://www.phila.gov/media/20160504172218/Green-Streets-Design-Manual-2014.pdf} \\
4 & Green Stormwater Infrastructure Maintenance Manual & 2016 & \url{https://water.phila.gov/wp-content/uploads/GSI-Maintenance-Manual_v2_2016.pdf} \\
5 & Green City, Clean Waters Plan & 2009 & \url{https://www.phila.gov/media/20160421133948/green-city-clean-waters.pdf} \\
6 & Green City, Clean Waters Partnership Agreement & 2012 & \url{https://water.phila.gov/wp-content/uploads/files/EPA_Partnership_Agreement.pdf} \\
7 & Green City Clean Waters - Long-Term Control Plan & -- & \url{https://water.phila.gov/wp-content/uploads/files/LTCPU_Complete.pdf} \\
8 & GSI Planning \& Design Manual & -- & \url{https://water.phila.gov/wp-content/uploads/files/gsi-planning-and-design-manual.pdf} \\
9 & GSI As-built Survey and Drafting Manual & -- & \url{https://water.phila.gov/wp-content/uploads/files/gsi-as-built-survey-and-drafting-manual.pdf} \\
10 & SMP Inspection Forms (Cisterns, Roofs, Ponds, Porous Surface, Basins, Filters) & -- &
\url{https://water.phila.gov/wp-content/uploads/files/smp-porous-surface-inspection-form.pdf}\\
11 & Philadelphia Water Department Regulations & 2024 & \url{https://water.phila.gov/wp-content/uploads/files/pwd-regulations-2024-04-29.pdf} \\
12 & Stormwater Grant Resources (portal) & -- & \url{https://water.phila.gov/stormwater/incentives/grants/} \\
13 & Plan and Report Checklists & -- & \url{https://water.phila.gov/wp-content/uploads/files/smgm-e-plan-and-report-checklists.pdf} \\
14 & Reported Flood Damages in Philadelphia (map) & 2024 & \url{https://www.phila.gov/media/20241204111812/Reported-Flood-Damages-Map-v4.2-2024.pdf} \\
15 & Sustainable Funding for Green City, Clean Waters & 2022 & \url{https://williampennfoundation.org/sites/default/files/2024-05/PHL-GreenCityCleanWaters-Sustain_2022_FINAL.pdf} \\
16 & GCCW Comprehensive Monitoring Plan & -- & \url{https://archive.phillywatersheds.org/ltcpu/GCCW%20Comprehensive%20Monitoring%20Plan%20Sections%201-10.pdf} \\
17 & PWD Regulations Chapter 6 - Stormwater & -- & \url{https://water.phila.gov/wp-content/uploads/files/pwd-regulations-chapter-6.pdf} \\
18 & SMP Maintenance Guidance & -- & \url{https://water.phila.gov/wp-content/uploads/files/smp-maintenance-guidance.pdf} \\
19 & SMP Maintenance Guide (portal) & -- & \url{https://water.phila.gov/development/stormwater-plan-review/maintenance} \\
20 & Rain Check Contractor Documents & -- & \url{https://www.pwdraincheck.org/en/contractor-documents} \\
21 & GSI Landscape Design Guidebook & 2014 & \url{https://www.pwdraincheck.org/images/documents/Landscape_Manual_2014.pdf} \\
22 & Planning \& Design Resource Directory (portal) & -- & \url{https://water.phila.gov/gsi/planning-design/resources/} \\
\bottomrule
\end{longtable}

\section{SFT Prompt Template}
\small
\begin{verbatim}
SYSTEM: You are an expert technical writer and dataset engineer for domain-specific LLM fine-tuning.

Your task is to read the provided PDF document and extract knowledge that is NOT generic LLM common knowledge, but instead is specific to this document, its policies, rules, responsibilities, procedures, or technical constraints.

You must generate a Supervised Fine-Tuning (SFT) dataset in JSON format.

IMPORTANT RULES (MUST FOLLOW):

1. Every QA pair MUST be fully self-contained and understandable WITHOUT access to the PDF.
   - Do NOT reference chapters, sections, figures, tables, or page numbers.
   - Do NOT use phrases like “Chapter 1”, “this section”, “as described above”, or “the following”.
   - Do NOT rely on document structure for meaning.

2. Avoid ambiguous pronouns and references.
   - DO NOT use: it, this, that, they, the City, the program, the agreement
   - INSTEAD, always explicitly name the entity:
     e.g., “the Philadelphia Water Department”, 
           “the Green City, Clean Waters program”, 
           “the municipal green stormwater infrastructure guidance”.

3. Each instruction must clearly state the domain and context.
   - A reader with no prior exposure to the PDF should still understand the question.
   - Example:
     Wrong: “Describe how pre-development land cover must be represented”
     Right: “Describe how pre-development land cover must be represented in stormwater modeling for municipal green stormwater infrastructure projects”

4. Each output must:
   - Be grounded ONLY in the PDF content
   - Restate key entities and constraints explicitly
   - Provide a clear, concrete, and technically meaningful answer
   - Be as long or as short as needed to fully capture the knowledge (no artificial length limits)

5. Each QA pair should represent ONE independent, reusable knowledge unit suitable for LLM fine-tuning.

6. If the document defines:
   - responsibilities → generate responsibility-focused QA
   - procedures → generate process-focused QA
   - design rules → generate constraint-focused QA
   - evaluation or monitoring → generate metric- or workflow-focused QA

OUTPUT FORMAT (STRICT):

[
  {
    "instruction": "A fully self-contained task or question",
    "input": "",
    "output": "A complete, document-grounded answer with explicit entities and no ambiguous references"
  }
]

Do NOT add explanations, commentary, or markdown.
Only output valid JSON.


USER PROMPT
Below is the extracted text content from a Green Stormwater Infrastructure (GSI) PDF document.

Your task:
1. Identify document-specific, non-obvious, and operationally relevant knowledge.
2. Convert that knowledge into self-contained instruction-style QA samples suitable for supervised fine-tuning (SFT).
3. Ensure that each question and answer can be fully understood without access to the original document.
4. Avoid vague entity references or pronouns unless the entity is explicitly defined in the instruction.

Generate as many high-quality samples as the content supports. Fewer high-quality samples are preferred over many weak ones.

Return ONLY a valid JSON array. Do not include explanations or markdown.

\end{verbatim}

% ------------------------
% Bibliography in the same file
% ------------------------

\bibliographystyle{plain}
\bibliography{refs}

\end{document}